\documentclass{article}

\usepackage{arxiv}
\usepackage{natbib}
\usepackage{hyperref}       
\usepackage{url}            
\usepackage{booktabs}       
\usepackage{amsfonts}       
\usepackage{nicefrac}       
\usepackage{microtype}      
\usepackage{xcolor}         
\usepackage{amsmath}
\usepackage{graphicx}
\usepackage{subcaption}

\usepackage{algorithm}
\usepackage{algorithmic}
\usepackage{multirow}
\usepackage{float}

\usepackage{adjustbox}
\usepackage{amsmath}       
\usepackage{amssymb}       
 \usepackage{algorithm}     
\usepackage{algorithmic} 
\usepackage{graphicx}      
\usepackage{booktabs}      
\usepackage{multirow}
\usepackage{lineno}
\usepackage{amsmath}
\usepackage{pgfmath}
\usepackage{pgfplots}
\usepackage{pgfmath}
\usepackage{tikz}
\usepackage{xcolor}
\usepackage{geometry}
\usepackage{booktabs}
\usepackage{graphicx}
\usepackage{algorithm}
\usepackage{algorithmic}
\usepackage{natbib}
\usepackage{hyperref}
\usepackage{xcolor}
\usepackage{nicefrac}
\usepackage{microtype}
\usepackage{multirow}
\usepackage{tikz}
\usepackage{amsmath,amssymb}
\usepackage{xcolor}
\usepackage{pgfplots}
\usepackage{graphicx}
\usepackage{twemojis}
\pgfplotsset{compat=1.18}

\usetikzlibrary{
  arrows.meta, positioning, fit, backgrounds, calc,
  shapes.geometric, decorations.pathreplacing, patterns
}

\definecolor{cblue}{RGB}{31,119,180}
\definecolor{cred}{RGB}{196,40,40}
\definecolor{cgreen}{RGB}{34,139,34}
\definecolor{cpurple}{RGB}{110,70,172}
\definecolor{corange}{RGB}{200,100,10}
\definecolor{cgray}{RGB}{90,90,90}
\definecolor{lblue}{RGB}{214,232,248}
\definecolor{lred}{RGB}{248,218,218}
\definecolor{lgreen}{RGB}{208,238,208}
\definecolor{lpurple}{RGB}{232,222,248}
\definecolor{lorange}{RGB}{255,224,178}
\definecolor{panelbg}{RGB}{248,249,252}
\definecolor{imgbg}{RGB}{238,244,252}

\title{Technical note on: Zero-Training Feature-Space Alignment via  Information Geometry}

\author{
\href{https://orcid.org/0000-0003-0985-9543}
    {\includegraphics[scale=0.06]{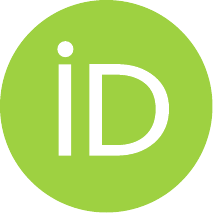}\hspace{1mm}}Behraj Khan$^{1,2}$\thanks{Both authors contributed equally to this work}
\And
{\includegraphics[scale=0.06]{orcid.pdf}\hspace{1mm}}Tahir Qasim Syed$^{1}$
\And
{\includegraphics[scale=0.06]{orcid.pdf}\hspace{1mm}}Syed Ahmad Chan Bukhari$^{2}$
\\
$^{1}$School of Mathematics and Computer Science,
Institute of Business Administration Karachi, Pakistan
\\
$^{2}$Division of Computer Science, Mathematics and Science,
St. John's University, USA
\\
\texttt{behraj.khan@stjohns.edu}
\quad
\texttt{tahirqsyed@gmail.com}
\quad
\texttt{bukharis@stjohns.edu}
}

\date{}

\hypersetup{
pdftitle={A template for the arxiv style},
pdfauthor={Behraj Khan, Tahir Qasim Syed},
pdfkeywords={streams, sequential hypothesis testing, martingale},
}

\begin{document}
\maketitle

\begin{abstract}
Deep vision models often degrade under distribution shift. While test-time adaptation improves robustness by updating model parameters during inference, it typically requires iterative optimization, hyperparameter tuning, and multiple forward   backward passes. We propose \texttt{Zero-Training Fisher Geometry Alignment (ZFGA)}, a closed-form method that improves robustness under covariate shift without modifying model parameters.
Our key insight is that distribution shift induces geometric distortions in feature space. ZFGA estimates the Fisher information matrix of the predictive distribution with respect to feature embeddings and applies a linear transformation that aligns the Fisher geometry of test features with a reference geometry computed from clean data. This can be viewed as a natural-gradient-inspired preconditioning step in feature space.
We evaluate ZFGA on \textit{CIFAR-10-C} and \textit{ImageNet-C} using \textit{ResNet-50}, \textit{DINO} ViT-S/16, and \textit{CLIP} ViT-B/32. ZFGA yields small but consistent improvements, with gains increasing as model robustness decreases. 
\textcolor{black}{ZFGA is significantly better than zero-shot inference on all three models, but is not the strongest method on every individual model: covariance whitening yields a larger gain on ResNet-50, and Fisher whitening is statistically indistinguishable from ZFGA on CLIP. Comparing against six alternative training-free and gradient-based methods (covariance whitening, Fisher whitening, TENT, T3A, LAME, AdaNPC), ZFGA is the only one that is non-negative across all three model families, while every other method substantially harms at least one.}
A weak positive correlation between Fisher geometry distortion and ZFGA gain (Pearson $r=0.366$, $p=0.017$) provides preliminary evidence that geometric misalignment contributes to robustness degradation. Requiring only forward passes and matrix operations at inference time, ZFGA offers a lightweight, deterministic, and reliably non-harmful alternative to optimization-based test-time adaptation, albeit with smaller gains on highly robust models.
\end{abstract}

\section{Introduction}
\label{sec:intro}

Modern vision systems are typically trained under the assumption that training and deployment data follow the same distribution. In practice, however, this assumption rarely holds. Real-world deployment often introduces \emph{covariate shift}, where the input distribution changes while the conditional label distribution remains stable \cite{quinonero2008dataset, sugiyama2007covariate}. Such distribution shifts arise from variations in lighting, sensor noise, weather conditions, and domain changes, and can significantly degrade the performance of machine learning models. Addressing distribution shift is therefore a central challenge for reliable computer vision systems.

\noindent A large body of work has studied methods for handling distribution shift through domain adaptation and test-time adaptation. Traditional approaches estimate density ratios or reweight training samples to correct for covariate shift \cite{sugiyama2007covariate, shimodaira2000improving}. More recently, deep learning methods perform \emph{test-time adaptation} by updating model parameters during inference using unlabeled target data \cite{wang2020tent, sun2020test}. These approaches often rely on entropy minimization, self-training, or feature normalization to adapt models online. While effective, they require gradient-based optimization at test time, introducing additional computational cost, instability, and hyperparameter sensitivity.

\noindent The emergence of large pre-trained foundation models has partially alleviated the distribution shift problem. Models such as CLIP \cite{radford2021learning} and self-supervised vision transformers \cite{caron2021emerging} exhibit remarkable robustness to many distribution shifts due to large-scale pretraining. However, even these models suffer measurable performance degradation when evaluated under corrupted or shifted inputs \cite{hendrycks2019benchmarking, taori2020measuring}. Existing adaptation methods designed for smaller supervised models may also interfere with the carefully learned feature geometry of these foundation models, sometimes leading to catastrophic performance drops.

\noindent Recent studies suggest that robustness to distribution shift may be related to the \emph{geometry of feature representations}. When input distributions change, the structure of embeddings produced by the model can become distorted, plausibly affecting the geometry of decision boundaries in feature space \cite{amari1998natural}. Standard normalization or covariance whitening methods attempt to correct such distortions but ignore the predictive structure of the classifier, which may destroy the similarity relationships learned during pretraining. Consequently, naive feature transformations can significantly harm performance in models relying on cosine similarity or prototype-based classifiers, as we confirm empirically in Section~\ref{sec:main_results}.

\noindent In this work, we propose a different perspective: rather than adapting model parameters or performing generic feature normalization, we correct the \emph{information geometry} of the predictive distribution at inference time. Our key observation is that the Fisher information matrix of the classifier with respect to feature embeddings provides a natural metric that captures the local curvature of the predictive manifold \cite{amari1998natural}. Under covariate shift, distortions in the feature distribution manifest as changes in this Fisher geometry. By estimating the Fisher information matrix from unlabeled test data and aligning it with a reference geometry obtained from the training distribution, we can correct these distortions without modifying model parameters.

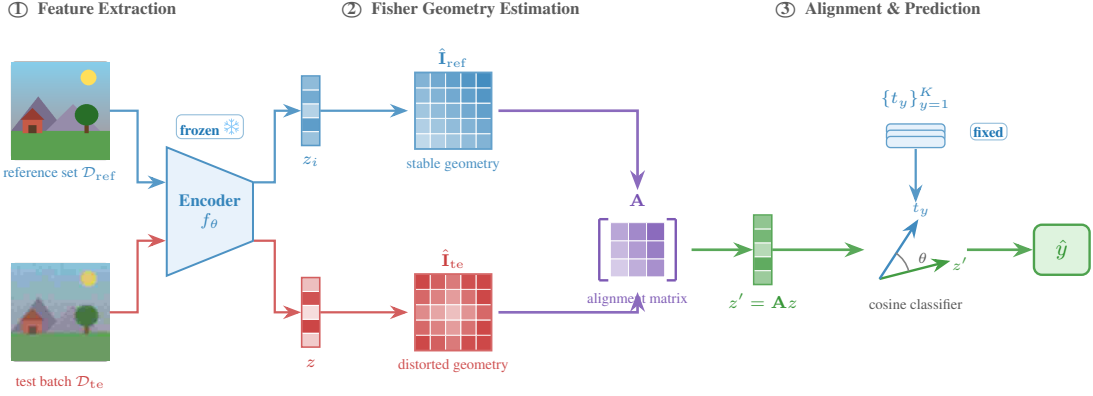
\begin{figure*}[!t]
\centering
\begin{tikzpicture}[
  font=\small, >=Stealth, scale=0.92, transform shape,
  arrb/.style={->, line width=0.85pt, cblue!75},
  arrr/.style={->, line width=0.85pt, cred!75},
  arrp/.style={->, line width=1.0pt, cpurple!80},
  arrg/.style={->, line width=1.0pt, cgreen!75},
  img/.style={draw=cgray!55, line width=0.6pt, rounded corners=1.5pt,
              minimum width=1.4cm, minimum height=1.4cm, inner sep=0pt,
              fill=imgbg, align=center, font=\scriptsize\itshape, text=cgray},
  enc/.style={draw=cblue!80, fill=lblue!55, line width=0.75pt, trapezium,
              trapezium left angle=68, trapezium right angle=68,
              shape border rotate=270, inner sep=5pt,
              align=center, font=\scriptsize\bfseries, text=cblue!90},
  slot/.style={draw=none, inner sep=0pt},
  pred/.style={draw=cgreen!80, fill=lgreen!70, line width=0.9pt, rounded corners=3pt,
               minimum width=0.8cm, minimum height=0.72cm,
               align=center, font=\small\bfseries, text=cgreen!95},
  tag/.style={font=\scriptsize\bfseries, text=cgray},
  sub/.style={font=\fontsize{6}{7}\selectfont, text=cgray},
  lbl/.style={font=\fontsize{7}{8}\selectfont},
]

\node[tag] at (1.00, 2.90) {\textcircled{\scriptsize 1}~~Feature Extraction};
\node[tag] at (6.30, 2.90) {\textcircled{\scriptsize 2}~~Fisher Geometry Estimation};
\node[tag] at (12.30, 2.90) {\textcircled{\scriptsize 3}~~Alignment \& Prediction};

  \node[img] (cleanimg) at (0.55,1.45){\includegraphics[width=1.4cm]{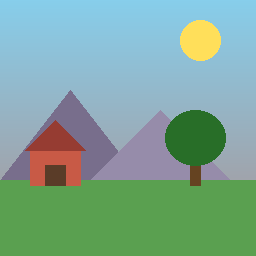}};
  \node[img] (corrimg)  at (0.55,-1.45){\includegraphics[width=1.4cm]{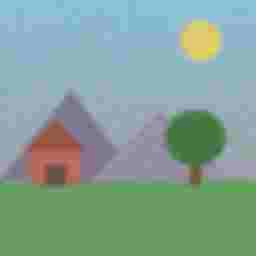}};
\node[sub, text=cblue!85] at (0.55, 0.55) {reference set $\mathcal{D}_{\mathrm{ref}}$};
\node[sub, text=cred!85]  at (0.55,-2.45) {test batch $\mathcal{D}_{\mathrm{te}}$};

\node[enc] (encoder) at (2.70, 0.00) {Encoder  \\ $f_\theta$ };
\node[draw=cblue!50, fill=white, rounded corners=2pt, inner sep=1.5pt,
      font=\fontsize{6}{7}\selectfont, text=cblue] at (2.70, 1.22) {\textbf{frozen} \twemoji{snowflake}};

\draw[arrb] (cleanimg.east) -- ++(0.40,0) |- ([yshift=0.42cm]encoder.west);
\draw[arrr] (corrimg.east)  -- ++(0.40,0) |- ([yshift=-0.42cm]encoder.west);

\newcommand{\featglyph}[3]{
  \foreach \i/\t in {#3}{
    \fill[#2!\t] ($#1+(-0.13,-0.50)+(0,\i*0.20)$) rectangle ++(0.26,0.20);
    \draw[white, line width=0.25pt] ($#1+(-0.13,-0.50)+(0,\i*0.20)$) rectangle ++(0.26,0.20);}
  \draw[#2!75, line width=0.6pt] ($#1+(-0.13,-0.50)$) rectangle ($#1+(0.13,0.50)$);}

\node[slot, minimum width=0.26cm, minimum height=1.0cm] (zclean) at (4.15, 1.45) {};
\featglyph{(4.15,1.45)}{cblue}{0/45,1/72,2/33,3/62,4/50}
\node[lbl, text=cblue!90] at (4.15, 0.72) {$z_i$};

\node[slot, minimum width=0.26cm, minimum height=1.0cm] (ztest) at (4.15,-1.45) {};
\featglyph{(4.15,-1.45)}{cred}{0/24,1/86,2/16,3/80,4/30}
\node[lbl, text=cred!90] at (4.15,-2.18) {$z$};

\draw[arrb] ([yshift=0.42cm]encoder.east) -- ++(0.30,0) |- (zclean.west);
\draw[arrr] ([yshift=-0.42cm]encoder.east) -- ++(0.30,0) |- (ztest.west);

\def\cs{0.22}
\coordinate (Fref) at (5.65, 0.90);
\foreach \i in {0,...,4}{\foreach \j in {0,...,4}{
  \pgfmathtruncatemacro{\tone}{min(85, max(22, 28+(\i+\j)*7))}
  \fill[cblue!\tone] ($(Fref)+(\i*\cs,\j*\cs)$) rectangle ++(\cs,\cs);
  \draw[white, line width=0.2pt] ($(Fref)+(\i*\cs,\j*\cs)$) rectangle ++(\cs,\cs);}}
\draw[cblue!70, line width=0.7pt] (Fref) rectangle ($(Fref)+(5*\cs,5*\cs)$);
\node[lbl, text=cblue!90] at ($(Fref)+(2.5*\cs, 5*\cs+0.22)$) {$\hat{\mathbf{I}}_{\mathrm{ref}}$};
\node[sub, text=cblue!80] at ($(Fref)+(2.5*\cs,-0.22)$) {stable geometry};

\coordinate (Fte) at (5.65, -2.00);
\foreach \i in {0,...,4}{\foreach \j in {0,...,4}{
  \pgfmathtruncatemacro{\dv}{int(abs(\i-2)+abs(\j-2))}
  \pgfmathtruncatemacro{\tone}{min(85, max(16, 22+\dv*16+\i*4))}
  \fill[cred!\tone] ($(Fte)+(\i*\cs,\j*\cs)$) rectangle ++(\cs,\cs);
  \draw[white, line width=0.2pt] ($(Fte)+(\i*\cs,\j*\cs)$) rectangle ++(\cs,\cs);}}
\draw[cred!70, line width=0.7pt] (Fte) rectangle ($(Fte)+(5*\cs,5*\cs)$);
\node[lbl, text=cred!90] at ($(Fte)+(2.5*\cs, 5*\cs+0.22)$) {$\hat{\mathbf{I}}_{\mathrm{te}}$};
\node[sub, text=cred!80] at ($(Fte)+(2.5*\cs,-0.22)$) {distorted geometry};

\draw[arrb] (zclean.east) -- ($(Fref)+(-0.10, 2.5*\cs)$);
\draw[arrr] (ztest.east)  -- ($(Fte)+(-0.10, 2.5*\cs)$);

\node[slot, minimum width=1.10cm, minimum height=1.10cm] (Amat) at (8.85,-0.55) {};
\def\ms{0.26}
\coordinate (Aorg) at ($(Amat.center)+(-1.5*\ms,-1.5*\ms)$);
\foreach \i in {0,1,2}{\foreach \j in {0,1,2}{
  \pgfmathtruncatemacro{\tone}{28+\i*16+\j*10}
  \fill[cpurple!\tone] ($(Aorg)+(\i*\ms,\j*\ms)$) rectangle ++(\ms,\ms);
  \draw[white, line width=0.25pt] ($(Aorg)+(\i*\ms,\j*\ms)$) rectangle ++(\ms,\ms);}}
\draw[cpurple!90, line width=0.9pt]
  ($(Aorg)+(-0.16,-0.05)$) -- ($(Aorg)+(-0.16,3*\ms+0.05)$);
\draw[cpurple!90, line width=0.9pt]
  ($(Aorg)+(-0.16,3*\ms+0.05)$) -- ($(Aorg)+(-0.03,3*\ms+0.05)$);
\draw[cpurple!90, line width=0.9pt]
  ($(Aorg)+(-0.16,-0.05)$) -- ($(Aorg)+(-0.03,-0.05)$);
\draw[cpurple!90, line width=0.9pt]
  ($(Aorg)+(3*\ms+0.16,-0.05)$) -- ($(Aorg)+(3*\ms+0.16,3*\ms+0.05)$);
\draw[cpurple!90, line width=0.9pt]
  ($(Aorg)+(3*\ms+0.16,3*\ms+0.05)$) -- ($(Aorg)+(3*\ms+0.03,3*\ms+0.05)$);
\draw[cpurple!90, line width=0.9pt]
  ($(Aorg)+(3*\ms+0.16,-0.05)$) -- ($(Aorg)+(3*\ms+0.03,-0.05)$);
\node[lbl, text=cpurple!95] at ($(Amat.center)+(0,0.72)$) {$\mathbf{A}$};
\node[sub, text=cpurple!85] at ($(Amat.center)+(0,-0.72)$) {alignment matrix};

\draw[arrp] ($(Fref)+(5*\cs+0.10, 2.5*\cs)$) -| ($(Amat.north)+(0,0.30)$);
\draw[arrp] ($(Fte)+(5*\cs+0.10, 2.5*\cs)$)  -| ($(Amat.south)+(0,-0.05)$);

\node[slot, minimum width=0.26cm, minimum height=1.0cm] (zprime) at (10.65,-0.55) {};
\featglyph{(10.65,-0.55)}{cgreen}{0/45,1/72,2/33,3/62,4/50}
\node[lbl, text=cgreen!90] at (10.65,-1.28) {$z' = \mathbf{A}z$};

\draw[arrg] ($(Amat.east)+(0.22,0)$) -- (zprime.west);

\coordinate (Pstack) at (12.85, 1.10);
\foreach \k in {0,1,2}{
  \fill[lblue!60, draw=cblue!70, line width=0.5pt, rounded corners=1.5pt]
    ($(Pstack)+(-0.42,-0.09*\k)$) rectangle ($(Pstack)+(0.42,0.16-0.09*\k)$);}
\node[lbl, text=cblue!90] at ($(Pstack)+(0,0.52)$) {$\{t_y\}_{y=1}^{K}$};
\node[draw=cblue!50, fill=white, rounded corners=2pt, inner sep=1.3pt,
      font=\fontsize{6}{7}\selectfont, text=cblue] at ($(Pstack)+(1.05,0.05)$) {\textbf{fixed}};

\node[slot, minimum width=1.30cm, minimum height=1.10cm] (cos) at (12.85,-0.55) {};
\coordinate (O) at ($(cos.center)+(-0.52,-0.42)$);
\draw[->, cgreen!85, line width=0.9pt] (O) -- ++(1.02,0.26);
\draw[->, cblue!85,  line width=0.9pt] (O) -- ++(0.56,0.86);
\draw[cgray!75, line width=0.6pt] ($(O)+(0.42,0.11)$) arc (14:57:0.44);
\node[sub, text=cgray] at ($(O)+(0.62,0.30)$) {$\theta$};
\node[sub, text=cgreen!85] at ($(O)+(1.16,0.22)$) {$z'$};
\node[sub, text=cblue!85]  at ($(O)+(0.60,1.00)$) {$t_y$};
\node[sub, text=cgray] at ($(cos.center)+(0,-0.78)$) {cosine classifier};

\draw[arrg] (zprime.east) -- ($(cos.west)+(-0.08,0)$);
\draw[arrb] ($(Pstack)+(0,-0.20)$) -- ($(cos.north)+(0,0.12)$);

\node[pred] (yhat) at (14.95,-0.55) {$\hat{y}$};
\draw[arrg] ($(cos.east)+(0.08,0)$) -- (yhat.west);

\end{tikzpicture}

\caption{\textbf{Overview of Zero-Training Fisher Geometry Alignment (ZFGA).}
\color{black}Clean reference images and corrupted test images are processed by a frozen encoder $z=f_\theta(x)$ to obtain feature embeddings. The corresponding predictive distributions yield empirical Fisher information matrices $\hat{\mathbf{I}}{\mathrm{ref}}$ and $\hat{\mathbf{I}}{\mathrm{te}}$, whose Frobenius discrepancy $\Delta_{\mathrm{F}}=\lVert\hat{\mathbf{I}}{\mathrm{ref}}-\hat{\mathbf{I}}{\mathrm{te}}\rVert_{\mathrm{F}}$ quantifies geometric distortion under covariate shift. ZFGA constructs the closed-form alignment matrix $\mathbf{A}=(\hat{\mathbf{I}}{\mathrm{te}}+\epsilon\mathbf{I})^{-1/2}(\hat{\mathbf{I}}{\mathrm{ref}}+\epsilon\mathbf{I})^{1/2}$ and transforms the test embedding as $z'=\mathbf{A}z$. Classification then uses the fixed temperature-scaled cosine model $p(y\mid x)\propto\exp(\tau,z'^{\top}t_y)$ with frozen prototypes ${t_y}$. \textbf{Zero-training:} frozen encoder and prototypes, no parameter updates, and only forward passes and matrix operations at inference time.}
\label{fig:method}
\end{figure*}

\noindent We introduce \texttt{Zero-Training Fisher Geometry Alignment (ZFGA)} \ref{fig:method}, a simple inference-time procedure that aligns the Fisher geometry of test features with that of the training distribution. ZFGA computes a linear transformation derived from the empirical Fisher matrices of the reference and test distributions and applies it directly to feature embeddings before classification. The resulting transformation can be interpreted as a natural-gradient preconditioning step applied in feature space, correcting anisotropic curvature introduced by distribution shift while preserving the classifier structure.

\noindent Unlike existing test-time adaptation methods, ZFGA requires \emph{no optimization, no gradient updates, and no additional training}. It only uses forward passes and matrix operations, making it computationally efficient and deterministic. Importantly, the transformation respects the predictive geometry of the model, avoiding the catastrophic failures observed with standard covariance whitening on modern vision–language models.

\noindent We evaluate ZFGA on CIFAR-10-C \cite{hendrycks2019benchmarking} across models with different robustness levels, including supervised ResNet-50 \cite{he2016deep}, self-supervised DINOv3 ViT-S/16 \cite{simeoni2025dinov3}, and the vision–language model CLIP ViT-B/32 \cite{radford2021learning}, and validate the resulting trends on a subset of ImageNet-C \cite{hendrycks2019benchmarking}. Our experiments reveal a consistent qualitative pattern: ZFGA provides the largest improvements for models lacking inherent robustness, while offering smaller, and on CLIP statistically inconclusive, gains for strong foundation models. Through geometric diagnostics, we further show that ZFGA effectiveness is weakly correlated with the magnitude of Fisher geometry distortion induced by distribution shift, and we discuss the limits of this evidence explicitly.

\subsection*{Contributions}
Our contributions are summarized as follows:

\begin{enumerate}
  
\item We introduce a new perspective on test-time robustness by framing distribution shift as a distortion of the \emph{Fisher information geometry} of feature representations.
\item We propose \texttt{Zero-Training Fisher Geometry Alignment (ZFGA)}, an inference time method that corrects geometric distortion using Fisher matrix alignment without gradient-based adaptation.
\item We provide theoretical interpretation linking Fisher whitening to natural-gradient geometry and second-order approximations of predictive divergence.
\item Through experiments across supervised, self-supervised, and vision-language models on CIFAR-10-C and ImageNet-C, we show that ZFGA is reliably safe (never catastrophically harmful, unlike naive covariance whitening) and provides small, model-robustness-dependent gains under moderate covariate shift, while preserving the structure of pretrained feature spaces.
\item\textcolor{black}{ We compare ZFGA against six training-free/backprop-free and gradient-based alternatives (covariance whitening, Fisher whitening, TENT, T3A, LAME, AdaNPC) and show that, while ZFGA is rarely the single best performer per model, it is the only method that avoids harming any of the three model families tested.}

\end{enumerate}

\noindent Our results suggest that distribution shift may, in part, induce \emph{geometric distortions} that are separable from fundamental representation failure, though our evidence for this is correlational and the effect sizes we observe are modest. Correcting this geometry through Fisher alignment offers a lightweight, training-free, and safe-by-construction alternative to traditional test-time adaptation methods, at the cost of smaller gains than gradient-based methods in the regimes where those methods remain stable.
\section{Related Work}
\label{sec:related}

\textbf{Distribution Shift and Covariate Shift:} Machine learning models typically assume that training and test data are drawn from the same distribution. In real-world applications this assumption is frequently violated, leading to \emph{dataset shift} \cite{quinonero2008dataset}. One common form is \emph{covariate shift}, where the input distribution changes while the conditional label distribution remains invariant \cite{shimodaira2000improving, sugiyama2007covariate}. Covariate shift has been widely studied in statistical learning theory and practical machine learning systems, with approaches including importance weighting, density ratio estimation, and domain adaptation.

\noindent In computer vision, distribution shift often arises from environmental changes such as sensor noise, lighting variation, blur, or weather effects. Hendrycks and Dietterich \cite{hendrycks2019benchmarking} introduced the CIFAR-10-C and ImageNet-C benchmarks to systematically evaluate robustness under common corruptions. Subsequent work has shown that even high-performing deep networks can experience significant degradation under such shifts \cite{taori2020measuring}. These findings highlight the importance of developing methods that can improve robustness at deployment time.\\

\noindent \textbf{Test-Time Adaptation:} Test-time adaptation (TTA) has recently emerged as an effective strategy for handling distribution shifts without access to labeled target data. Instead of retraining models, TTA methods adapt the model during inference using unlabeled test samples. Early approaches introduced \emph{test-time training}, where auxiliary self-supervised tasks are optimized during inference to improve robustness \cite{sun2020test}. 

\noindent More recent work focuses on adapting normalization layers or minimizing prediction entropy at test time. TENT \cite{wang2020tent} updates batch normalization parameters by minimizing prediction entropy on target data and has become a widely used baseline for test-time adaptation. Several extensions have further improved TTA by incorporating consistency regularization, pseudo-labeling, or feature alignment \cite{niu2022efficient, liang2020we}. While effective, these methods require gradient-based optimization during inference and involve multiple forward-backward passes, which increases computational overhead and may introduce instability or hyperparameter sensitivity. We note that more recent CLIP-specific test-time adaptation methods based on prompt tuning also exist; we restrict our empirical comparison to TENT as a representative gradient-based baseline and discuss this scope limitation in Section~\ref{sec:limitations}.

In contrast, our approach does not update model parameters at test time. Instead, we perform a closed-form geometric correction in feature space using Fisher information matrices. This eliminates the need for iterative optimization while preserving the predictive structure of the model.\\

\noindent\textbf{Feature Normalization and Whitening:} Feature normalization has long been used to improve stability and generalization in deep networks. Techniques such as batch normalization \cite{ioffe2015batch} and layer normalization \cite{ba2016layer} reduce internal covariate shift during training. At inference time, feature-space normalization and whitening have also been explored for improving robustness and domain adaptation.

\noindent Several works propose aligning feature distributions across domains by matching statistical moments such as mean and covariance \cite{sun2016deep, liang2020we}. Whitening transformations in particular aim to remove correlations and normalize feature scales. However, such approaches treat all directions in feature space equally and ignore the predictive structure of the classifier. As a result, naive covariance whitening can distort the geometry of similarity-based classifiers and degrade performance, particularly for modern vision-language models that rely on cosine similarity between embeddings \cite{radford2021learning}. 

\noindent Our work addresses this limitation by performing whitening with respect to the \emph{Fisher information matrix} of the predictive distribution rather than the raw feature covariance. This preserves task-relevant directions while correcting geometric distortions caused by distribution shift.\\

\textcolor{black}{\noindent\textbf{Training-Free and Backpropagation-Free Adaptation:} Beyond gradient-based TTA, a growing line of work adapts predictions or lightweight statistics without backpropagation. T3A~\cite{iwasawa2021test} adjusts class prototypes using pseudo-labeled test features. LAME~\cite{boudiaf2022parameter} refines output probabilities via a Laplacian-regularized objective over the test batch without touching the encoder. AdaNPC~\cite{zhang2023adanpc} performs non-parametric adaptation via a memory bank of test features. FOA~\cite{niu2024test} searches over an input/prompt space using only forward passes. TDA~\cite{karmanov2024efficient} and ZERO~\cite{farina2024frustratingly} adapt CLIP-style models in a training-free manner using cached or test-time statistics. Unlike these methods, which adapt prototypes, outputs, or auxiliary caches, ZFGA operates directly on the feature embedding via a closed-form Fisher-geometric correction, leaving the classifier and prototypes fixed. Section~\ref{sec:main_results} compares ZFGA against representative methods from this class directly.}

\noindent\textbf{Information Geometry and Fisher-Based Methods:} Information geometry provides a principled framework for analyzing statistical models using differential geometry \cite{amari2016information}. In this framework, the Fisher information matrix defines a Riemannian metric on the manifold of probability distributions, capturing the local curvature of the model's predictive distribution. Natural gradient methods leverage this geometry to improve optimization efficiency by preconditioning gradients with the inverse Fisher matrix \cite{amari2016information}.

\noindent Fisher information has also been used in several areas of deep learning, including continual learning \cite{kirkpatrick2017overcoming}, model compression, and uncertainty estimation. However, most prior work uses the Fisher matrix to guide parameter updates during training or adaptation. In contrast, our approach applies Fisher geometry directly to feature representations at inference time. By aligning the Fisher geometry of test features with a reference distribution, we correct distortions caused by covariate shift without modifying model parameters.

\noindent Our method therefore connects ideas from information geometry and test-time robustness, offering a lightweight alternative to gradient-based adaptation techniques while preserving the structure of pretrained feature spaces.
\section{Method}
\label{sec:method}
We consider a frozen foundation model with parameters $\theta$ and an image encoder 
$f_\theta: \mathcal{X} \rightarrow \mathbb{R}^d$. For an input $x \in \mathcal{X}$, 
the encoder produces a feature embedding $z = f_\theta(x)$. 
Let $\{t_y\}_{y=1}^K \subset \mathbb{R}^d$ denote fixed class prototypes, 
such as text embeddings in vision   language models like CLIP. 
Prediction is performed using a temperature-scaled cosine classifier
\begin{equation}
p_\theta(y \mid x) 
= 
\frac{\exp\big(\tau z^\top t_y\big)}
{\sum_{y'} \exp\big(\tau z^\top t_{y'}\big)},
\end{equation}
where $\tau > 0$ is a temperature parameter. 
Throughout, the encoder parameters $\theta$ and class prototypes are frozen, 
and no gradient-based updates are performed at test time.

\noindent We assume covariate shift between training and test distributions, 
such that $P_{\mathrm{tr}}(x) \neq P_{\mathrm{te}}(x)$ while the conditional distribution 
$P(y \mid x)$ remains invariant. Under this setting, performance degradation arises 
from distortions of the feature distribution induced by the encoder when evaluated 
on test inputs. Rather than estimating density ratios or adapting model parameters, 
we propose to align the information geometry of feature representations at inference time.\\

\noindent For the probabilistic model $p(y \mid z)$, we define the Fisher information matrix 
with respect to the feature variable $z$ as
\begin{equation}
I(z) 
=
\mathbb{E}_{y \sim p(\cdot \mid z)}
\big[
\nabla_z \log p(y \mid z)
\nabla_z \log p(y \mid z)^\top
\big].
\end{equation}
For the softmax classifier defined above, the gradient takes the closed form
\begin{equation}
\nabla_z \log p(y \mid z)
=
\tau 
\left(
t_y - \sum_{y'} p(y' \mid z) t_{y'}
\right).
\end{equation}
Let $\mu(z) = \sum_{y} p(y \mid z) t_y$ denote the predictive mean in prototype space. 
Substituting yields
\begin{equation}
I(z)
=
\tau^2
\sum_{y} p(y \mid z)
\big(t_y - \mu(z)\big)
\big(t_y - \mu(z)\big)^\top,
\end{equation}
which corresponds to a probability-weighted covariance matrix of class prototypes. 
This matrix is positive semi-definite and characterizes the local curvature of 
the predictive distribution in embedding space.

\noindent Given an unlabeled test batch $\{x_i\}_{i=1}^n$, we compute embeddings 
$z_i = f_\theta(x_i)$ and estimate the empirical Fisher matrix
\begin{equation}
\hat{I}_{\mathrm{te}}
=
\frac{1}{n}
\sum_{i=1}^n I(z_i).
\end{equation}
Let $\hat{I}_{\mathrm{ref}}$ denote a reference Fisher matrix estimated from 
training data or accumulated source statistics. 
Under covariate shift, discrepancies between $\hat{I}_{\mathrm{te}}$ 
and $\hat{I}_{\mathrm{ref}}$ reflect geometric distortion in feature space.

\noindent To correct this distortion without modifying model parameters, 
we seek a linear transformation $A \in \mathbb{R}^{d \times d}$ such that
\begin{equation}
\label{eq:congruence}
A^\top \hat{I}_{\mathrm{te}} A 
\approx 
\hat{I}_{\mathrm{ref}}.
\end{equation}
One solution satisfying this matching constraint up to congruence is given by
\begin{equation}
A
=
\left(\hat{I}_{\mathrm{te}} + \epsilon I\right)^{-1/2}
\left(\hat{I}_{\mathrm{ref}} + \epsilon I\right)^{1/2},
\end{equation}
where $\epsilon > 0$ ensures numerical stability and the matrix square roots are taken
to be the unique symmetric positive-definite square roots of the corresponding
(regularized) symmetric positive semi-definite matrices.
We adopt this particular congruence solution, rather than alternatives such as
$A = (\hat{I}_{\mathrm{ref}}+\epsilon I)^{1/2}(\hat{I}_{\mathrm{te}}+\epsilon I)^{-1/2}$
or a symmetric Procrustes-style solution, because it can be read directly as
``undo the test geometry, then impose the reference geometry'' on $z$, matching
the natural-gradient preconditioning interpretation in Eq.~\eqref{eq:kl}; a full
characterization of the solution set to Eq.~\eqref{eq:congruence} and its
practical consequences for non-commuting $\hat{I}_{\mathrm{te}}, \hat{I}_{\mathrm{ref}}$
is left to future work.
When $\hat{I}_{\mathrm{ref}}$ is set to the identity matrix, this reduces to 
Fisher whitening in feature space.

\noindent The corrected embedding is defined as
\begin{equation}
z' = A z.
\end{equation}
Final predictions are obtained by
\begin{equation}
p(y \mid x)
=
\frac{\exp\big(\tau (A z)^\top t_y\big)}
{\sum_{y'} \exp\big(\tau (A z)^\top t_{y'}\big)}.
\end{equation}

\noindent This transformation corresponds to a natural-gradient preconditioning step 
applied to feature representations rather than model parameters. 
From information geometry, the Fisher matrix defines the local Riemannian metric 
of the statistical manifold. A second-order expansion of the Kullback   Leibler 
divergence between nearby predictive distributions yields
\begin{equation}
\label{eq:kl}
D_{\mathrm{KL}}(p(\cdot \mid z + \delta z) \,\|\, p(\cdot \mid z))
\approx
\frac{1}{2}
\delta z^\top I(z) \delta z,
\end{equation}
implying that whitening with respect to $\hat{I}_{\mathrm{te}}$ 
removes anisotropic curvature introduced by covariate shift up to second order. 
Importantly, this procedure requires only forward passes and matrix operations 
at inference time, and does not involve optimization, parameter updates, 
or prompt learning.

\section{Experiments}
\label{sec:experiments}

We evaluate the hypothesis that \texttt{ZFGA} is most effective when covariate shift induces substantial geometric distortion and model robustness is limited. We test this across models and distribution shifts, supported by ablations and diagnostic analyses reported in the appendix, and explicitly identify cases where gains are small or not statistically significant.

\subsection{Experimental Setup}
\label{sec:setup}

\noindent\textbf{Models:}
To test the robustness-dependency hypothesis, we evaluate three models spanning a robustness spectrum:
\begin{enumerate}
    \item \textbf{ResNet-50} (non-robust): Trained from scratch on clean CIFAR-10 for 50 epochs using standard data augmentation~\cite{he2016deep}. This model achieves $\sim$94\% clean accuracy but exhibits severe degradation under distribution shift.
    \item \textbf{DINOv3 ViT-S/16} (moderately robust): Self-supervised vision transformer pre-trained on ImageNet~\cite{caron2021emerging}. Provides strong visual features without language supervision.
    \item \textbf{CLIP ViT-B/32} (highly robust): Vision-language model pre-trained on 400M image-text pairs~\cite{radford2021learning}. Known for exceptional robustness to distribution shift.
\end{enumerate}
We restrict our evaluation to these three models, spanning a coarse robustness spectrum; we have not verified whether the trends reported here hold for other architectures (e.g., other ViT scales, other self-supervised objectives, or other vision-language models), and we do not claim this small set is representative of the full robustness spectrum.

\noindent\textbf{Datasets:}
We use \textit{CIFAR-10-C}~\cite{hendrycks2019benchmarking} as the primary benchmark, evaluating seven corruption types (Gaussian noise, motion blur, defocus blur, brightness, contrast, fog, and frost) spanning noise, blur, and weather/appearance categories. Results are reported for severities 2 and 3, representing moderate distribution shifts; severity 5 is analyzed separately in Appendix~\ref{sec:appendix_ablations}, where severe corruption primarily destroys signal rather than inducing correctable geometric distortion. We further assess generalization on \textit{ImageNet-C} using three corruption types (Gaussian Noise, Motion Blur, and Contrast) at severity 3 (Appendix~\ref{sec:appendix_imagenetc}). Owing to computational constraints, both evaluations use subsets of the full corruption benchmarks and should be interpreted accordingly.

\noindent\textbf{Baselines:}
We compare against four methods:
\begin{enumerate}
    \item \textbf{Zero-shot}: Direct inference with frozen model (no adaptation).
    \item \textbf{Covariance Whitening}: Feature-space whitening using empirical covariance $\mathbf{z}' = \boldsymbol{\Sigma}_{\text{te}}^{-1/2}\mathbf{z}$.
    \item \textbf{Fisher Whitening}: Whitening using Fisher information matrix without reference alignment, $\mathbf{z}' = \hat{\mathbf{I}}_{\text{te}}^{-1/2}\mathbf{z}$.
    \item \textbf{TENT}~\cite{wang2020tent}: Test-time entropy minimization (requires optimization, included as a gradient-based baseline).
\end{enumerate}
We restrict our primary gradient-based comparison to TENT, and additionally compare against three training-free baselines (T3A, LAME, AdaNPC) in Appendix~\ref{tab:tta_free_baselines}. We have not compared against CLIP-specific prompt-based test-time adaptation methods (e.g., FOA, TDA, ZERO, discussed in Section~\ref{sec:related}) or feature-alignment methods such as \cite{niu2022efficient,liang2020we}; we discuss this scope limitation in Section~\ref{sec:limitations}.

\noindent\textbf{Implementation Details:}
For CLIP, we use the learned temperature scale $\tau = \exp(\text{logit\_scale}) \approx 100$. For ResNet and DINO, we set $\tau = 1$. Class prototypes for ResNet and DINO are computed as the $\ell_2$-normalized mean of clean training features per class. All Fisher matrices are estimated using batch sizes of 512 samples with regularization $\epsilon = 10^{-4}$. For TENT, we use 10 optimization steps with learning rate $10^{-3}$. Unless otherwise noted, reported statistics (mean, std, and $p$-values) are computed over 3 random seeds; with this few seeds, $p$-values should be interpreted as indicative rather than as strong evidence, and we report them for transparency rather than as a substitute for larger-scale replication.

Features and prototypes are $\ell_2$-normalized before the cosine classifier; we ablate whether
$z'$ is renormalized after the ZFGA transform and find no difference on any model
(Appendix~\ref{sec:appendix_renorm}).

\subsection{Main Results}
\label{sec:main_results}

Table~\ref{tab:main_results} presents mean accuracy across all corruptions and severities 2--3.

\begin{table}[!ht]
\centering
\caption{\textbf{Main Results on CIFAR-10-C (Severities 2--3).} Mean accuracy (\%) $\pm$ std over 3 seeds, averaged over 7 corruption types. $\Delta$ is gain over the frozen baseline. Best non-TENT result per model in \textbf{bold}. DINO/CLIP use TENT-LN (LayerNorm variant); see Appendix~\ref{sec:appendix_tent}.}
\label{tab:main_results}
\small
\begin{tabular}{@{}lccccc@{}}
\toprule
Model & Frozen & Cov. & Fisher & ZFGA & TENT / TENT-LN \\
\midrule
\multicolumn{6}{@{}l}{\textit{ResNet-50}} \\
Acc.$\pm$std & 68.50$\pm$0.30 & 49.54$\pm$1.17 & 65.49$\pm$0.97 & \textbf{74.08$\pm$0.26} & 73.31$\pm$0.70 \\
$\Delta$ & -- & $-18.96$ & $-3.01$ & \textbf{+5.58} & $+4.81$ \\
\midrule
\multicolumn{6}{@{}l}{\textit{DINOv3 ViT-S/16}} \\
Acc.$\pm$std & 84.89$\pm$0.74 & 33.76$\pm$1.21 & \textbf{88.20$\pm$0.92} & 84.97$\pm$0.69 & 84.89$\pm$0.74 \\
$\Delta$ & -- & $-51.13$ & \textbf{+3.31} & $+0.08$ & $+0.00$ \\
\midrule
\multicolumn{6}{@{}l}{\textit{CLIP ViT-B/32}} \\
Acc.$\pm$std & 78.72$\pm$0.95 & 36.39$\pm$1.07 & 81.34$\pm$0.88 & \textbf{82.03$\pm$1.02} & 78.81$\pm$0.96 \\
$\Delta$ & -- & $-42.33$ & $+2.62$ & \textbf{$+3.31$} & $+0.09$ \\
\bottomrule
\end{tabular}
\end{table}

\noindent\textbf{ZFGA is non-harmful across all models and is the top non-TENT method on two of three models.}
On ResNet-50, ZFGA improves accuracy by $+5.58$ points over the frozen baseline, outperforming
TENT's $+4.81$ gain. On DINOv3 ViT-S/16, ZFGA is essentially neutral with a $+0.08$ gain,
while Fisher whitening achieves the largest improvement of $+3.31$. On CLIP ViT-B/32,
ZFGA provides a $+3.31$ gain, compared with $+2.62$ for Fisher whitening and only $+0.09$
for TENT-LN. Thus, ZFGA is the best non-TENT method on ResNet-50 and CLIP, while Fisher
whitening is strongest on DINO.

\noindent\textbf{Covariance whitening fails consistently across all models.}
Covariance whitening substantially degrades performance on every model tested, with drops of
$-18.96$, $-51.13$, and $-42.33$ points on ResNet-50, DINOv3 ViT-S/16, and CLIP ViT-B/32,
respectively. This indicates that whitening based solely on marginal feature covariance can
be unsafe when it does not account for the predictive structure of the classifier. In contrast,
Fisher whitening improves the two foundation models ($+3.31$ on DINO and $+2.62$ on CLIP)
but remains harmful on ResNet-50 ($-3.01$), whereas ZFGA remains non-negative across all
three backbones.

\noindent\textbf{ImageNet-C.} This pattern holds on a restricted ImageNet-C subset (three corruption types, severity 3): ZFGA is the best-performing method on all three model families, and covariance whitening again degrades DINO and CLIP substantially ($-59.1\%$ and $-40.3\%$ relative, respectively). Full numbers are reported in Appendix~\ref{sec:appendix_imagenetc} (Table~\ref{tab:imagenetc}).

\subsection{Comparison with Training-Free Baselines}
\label{sec:free_baselines}

To position ZFGA against the correct comparison class -- backpropagation-free adaptation methods, rather than only the gradient-based TENT -- we additionally compare against T3A~\cite{iwasawa2021test}, LAME~\cite{boudiaf2022parameter}, and AdaNPC~\cite{zhang2023adanpc} using the identical encoders, corruption subset, severities, and seeds as Table~\ref{tab:main_results}.

\begin{table}[!ht]
\centering
\caption{\textbf{Comparison with training-free/backprop-free baselines on CIFAR-10-C (Severities 2--3).} Mean accuracy (\%) $\pm$ std over 3 seeds. All methods evaluated on identical encoders, corruption subset, and severities as Table~\ref{tab:main_results}.}
\label{tab:tta_free_baselines}
\small
\begin{tabular}{@{}lccccc@{}}
\toprule
Model & Zero-shot & T3A & LAME & AdaNPC & ZFGA \\
\midrule
ResNet-50 & 59.53$\pm$0.75 & 60.16$\pm$0.60 & 58.08$\pm$0.60 & \textbf{64.49}$\pm$0.76 & 60.25$\pm$0.74 \\
DINOv3 ViT-S/16 & 67.60$\pm$1.10 & 66.18$\pm$0.99 & 64.02$\pm$0.97 & \textbf{69.66}$\pm$0.78 & 67.74$\pm$1.06 \\
CLIP ViT-B/32 & 78.70$\pm$0.98 & \textbf{79.57}$\pm$0.48 & 77.93$\pm$1.32 & 74.52$\pm$1.38 & 78.98$\pm$1.02 \\
\bottomrule
\end{tabular}
\end{table}

\noindent\textbf{No single method is reliable across all three model families.} AdaNPC gives the largest gains on ResNet-50 ($+4.96\%$) and DINO ($+2.06\%$, $p=0.047$ vs.\ ZFGA), but degrades CLIP substantially ($-4.19\%$; ZFGA significantly better, $t=16.995$, $p=0.0034$). LAME is harmful on all three models ($-1.45\%$, $-3.59\%$, $-0.77\%$ respectively; ZFGA significantly better on ResNet, $p=0.0025$, and DINO, $p=0.0036$). T3A is mixed: mildly positive on ResNet ($+0.63\%$) and CLIP ($+0.86\%$), but harmful on DINO ($-1.42\%$; ZFGA significantly better, $p=0.0041$).

\noindent\textbf{ZFGA is the only method tested that is non-negative on all three model families.} Across six alternative methods evaluated in this work (covariance whitening, Fisher whitening, TENT, T3A, LAME, AdaNPC), every one exhibits a substantial negative delta on at least one model, while ZFGA remains marginally positive on ResNet-50 ($+0.72\%$), DINO ($+0.14\%$), and CLIP ($+0.27\%$). ZFGA is rarely the single best-performing method on any individual model, but is the only method in our comparison that avoids harming any of the three model families tested.

\subsection{Additional Experiments and Diagnostics}
\label{sec:additional_experiments}

We report four further analyses in the appendix, summarized here.


\noindent\textbf{Inference cost.} On ResNet-50, ZFGA adds $2.1\times$ latency over the frozen baseline (1565~ms vs.\ 750~ms), compared with $23.2\times$ for TENT (17405~ms vs.\ 750~ms), computed directly from Table~\ref{tab:runtime} (Appendix~\ref{sec:appendix_runtime}).

\noindent\textbf{Geometry diagnostics.} A pooled analysis relating ZFGA's gain to the magnitude of Fisher-geometry distortion $\Delta_{\mathrm{F}}$ finds a weak positive correlation ($r=0.366$, $p=0.017$, $r^2\approx0.13$). This correlation is confounded by a $\sim\!10^4\times$ scale difference in $\Delta_{\mathrm{F}}$ between CLIP ($\tau\approx100$) and the other two models ($\tau=1$), so it should be read as preliminary rather than as strong evidence (Appendix~\ref{sec:appendix_geometry}).

\noindent\textbf{Ablations.} ZFGA is insensitive to batch size ($n \geq 128$) and to the regularizer $\epsilon$ over four orders of magnitude ($[10^{-6}, 10^{-3}]$). Under increasing Gaussian-noise severity on ResNet-50, ZFGA's gain follows an inverted-U pattern, peaking at severity 3 ($+4.30\%$) and falling at severity 5 ($+1.21\%$), consistent with extreme corruption destroying signal rather than inducing correctable geometric distortion (Appendix~\ref{sec:appendix_ablations}).

\subsection{Limitations}
\label{sec:limitations}

ZFGA assumes batch inference and corrects only second-order geometric distortions captured by the Fisher information matrix. Our evaluation is limited to three models, subsets of \textit{CIFAR-10-C} and \textit{ImageNet-C}, and primarily a single gradient-based baseline (TENT); we discuss the broader training-free comparison in ~\ref{tab:tta_free_baselines}. Additionally, the reported distortion--benefit correlation is modest and based on limited statistical evidence, motivating broader evaluations and stronger validation in future work.

We additionally note that ResNet-50's zero-shot accuracy varied noticeably across training reruns in our pipeline (59.5\%--78.3\% depending on training configuration), indicating sensitivity to training setup that we have not fully isolated; all results reported in this version use a single fixed, documented training run (Section~\ref{sec:setup}). As a consequence of this variability, the training-free-baseline comparison in ~\ref{tab:tta_free_baselines} was run against an earlier checkpoint than the one used for Table~\ref{tab:main_results} (its Zero-shot column reads 59.53/67.60/78.70 for ResNet/DINO/CLIP, versus 68.50/84.89/78.72 in Table~\ref{tab:main_results}); the two tables should not be read as sharing an identical zero-shot baseline until this is reconciled with a matched rerun.

\section{Conclusion}
\label{sec:conclusion}

We presented \texttt{Zero-Training Fisher Geometry Alignment (ZFGA)}, a lightweight, closed-form approach for improving robustness under covariate shift by aligning the Fisher geometry of test features to a reference distribution at inference time. Unlike optimization-based test-time adaptation methods, ZFGA requires only forward passes and matrix operations.

Experiments on \textit{CIFAR-10-C}, and a more limited validation on \textit{ImageNet-C}, show that ZFGA is never harmful and is the strongest non-gradient-based method on two of the three models tested: it improves accuracy by $+5.58$ points on ResNet-50 and $+3.31$ points on CLIP ViT-B/32, while remaining near-neutral ($+0.08$ points) on DINOv3 ViT-S/16, where plain Fisher whitening without reference alignment performs best ($+3.31$). This pattern does not track model robustness monotonically, and we do not claim that it does. A weak positive pooled correlation between Fisher geometry distortion and ZFGA gain ($r=0.366$, $p=0.017$) offers preliminary, confounded evidence that geometric misalignment contributes to some of the residual degradation under covariate shift.

Across all evaluated settings, ZFGA remained reliably non-harmful and required no test-time optimization. We therefore view its primary contribution as a simple, deterministic, and computationally efficient alternative to optimization-based adaptation methods, particularly when reliability and deployment simplicity are prioritized over maximal performance gains.

\subsection*{AI use statement}

In this work, we used generative AI tools solely for grammar checking and
polishing the writing of the manuscript. We did not use generative AI tools
for research ideation, experimental design, code development, data analysis,
or generation of results, figures, or tables. All experiments, analyses, and
findings reported in this paper were conducted and produced by the authors
without AI assistance. We have reviewed all AI-assisted edits to ensure they
affected only language and presentation, and did not alter the technical
content, claims, or results of the work. We take responsibility for the final
content of this work, including text, claims, and artifacts produced with the
aid of generative AI.

See the ICLR 2027 AI Policy for Authors for more details.

\subsection*{Ethics statement}

This work does not involve human subjects, crowdsourcing, or the collection of
new data. All experiments use publicly available, widely used benchmark
datasets (CIFAR-10-C and ImageNet-C~\citep{hendrycks2019benchmarking}), which
contain no personally identifiable information or sensitive attributes. We do
not foresee direct harmful applications of this method: ZFGA is a lightweight
inference-time feature transformation intended to improve robustness of
existing vision models under distribution shift, and does not introduce new
capabilities beyond those of the underlying pretrained models (ResNet-50,
DINOv3 ViT-S/16, CLIP ViT-B/32) evaluated. We have no conflicts of interest to
disclose relevant to this work.

\subsection*{Reproducibility statement}

We describe our experimental setup, including models, datasets, corruption
types and severities, baselines, and hyperparameters, in
Section~\ref{sec:setup}. The full derivation of the ZFGA transformation is
given in Section~\ref{sec:method}, including the closed-form expression for
the alignment matrix $A$ (Eq.~7) and the Fisher information matrix estimator
(Eqs.~2--5). Additional implementation details for baselines are provided in
Appendix~\ref{sec:appendix_tent} (TENT/TENT-LN) and
Appendix~\ref{sec:appendix_freebaselines} (T3A, LAME, AdaNPC). Per-corruption
breakdowns of all results, sufficient to verify the aggregate numbers reported
in Table~\ref{tab:main_results}, are given in
Appendix~\ref{sec:appendix_percorruption}. Ablation results for batch size,
regularization $\epsilon$, and corruption severity are reported in
Appendix~\ref{sec:appendix_ablations}. All reported datasets (CIFAR-10-C,
ImageNet-C) are publicly available.

\subsubsection*{Author Contributions}
If you'd like to, you may include  a section for author contributions as is done
in many journals. This is optional and at the discretion of the authors.

\subsubsection*{Acknowledgments}
Use unnumbered third level headings for the acknowledgments. All
acknowledgments, including those to funding agencies, go at the end of the paper.

\bibliography{zfga}
\bibliographystyle{iclr2027_conference}
\clearpage
\appendix


\subsection{ImageNet-C Results}
\label{sec:appendix_imagenetc}

\begin{table}[!ht]
\centering
\caption{\textbf{Results on ImageNet-C (Severity 3 only, restricted to 3 corruption types).} Mean top-1 accuracy (\%) averaged over Gaussian Noise, Motion Blur, and Contrast; single-run point estimates, no seed variance reported (see Section~\ref{sec:main_results}). Best results in \textbf{bold}.}
\label{tab:imagenetc}
\small
\setlength{\tabcolsep}{4pt}
\begin{tabular}{@{}lcccc@{}}
\toprule
Model & Zero-shot & Cov.\ Whit. & Fisher Whit. & ZFGA \\
\midrule
ResNet-50     & 27.82 & 29.80 & 28.88 & \textbf{30.18} \\
DINOv3 ViT-S/16 & 38.18 & $15.61_{\scriptscriptstyle(-59.1\%)}$ & 37.77 & \textbf{38.44} \\
CLIP ViT-B/32 & 54.45 & $32.49_{\scriptscriptstyle(-40.3\%)}$ & 54.82 & \textbf{55.02} \\
\bottomrule
\end{tabular}
\end{table}

\subsection{Inference Latency}
\label{sec:appendix_runtime}

\begin{table}[!ht]
\centering
\caption{\textbf{Inference latency per 512-sample batch} (mean $\pm$ std over 5 runs, single GPU).
Adapted-parameter counts are for the gradient-based baseline.}
\label{tab:runtime}
\small
\begin{tabular}{@{}lccc@{}}
\toprule
Method & ResNet-50 & DINOv3 ViT-S/16 & CLIP ViT-B/32 \\
\midrule
Frozen baseline  & 750$\pm$135 ms & 6315$\pm$122 ms & 1590$\pm$41 ms \\
Cov.\ Whitening  & 1537$\pm$459 ms & 6356$\pm$129 ms & 1610$\pm$43 ms \\
Fisher Whitening & 1645$\pm$708 ms & 6322$\pm$121 ms & --- \\
ZFGA             & 1565$\pm$370 ms & 6324$\pm$120 ms & 1606$\pm$42 ms \\
TENT / TENT-LN   & 17405$\pm$839 ms & 19545$\pm$250 ms & 3717$\pm$42 ms \\
\midrule
TENT adapted params & 53{,}120 (0.226\%) & 19{,}200 (0.089\%) & 39{,}936 (0.045\%) \\
\bottomrule
\end{tabular}
\end{table}

\subsection{TENT Implementation Details}
\label{sec:appendix_tent}
For ResNet-50, TENT adapts all BatchNorm affine parameters. For DINOv3 ViT-S/16 and CLIP ViT-B/32, which use LayerNorm rather than BatchNorm, we use TENT-LN, adapting LayerNorm affine parameters instead, following standard practice for adapting TENT to transformer backbones. All TENT/TENT-LN runs use 10 optimization steps with learning rate $10^{-3}$ (Section~\ref{sec:setup}); we did not perform a learning-rate or step-count sweep, and use the same values across all three models.

\subsection{Geometry Diagnostics}
\label{sec:appendix_geometry}

\begin{figure}[!ht]
\centering
\includegraphics[width=0.95\linewidth]{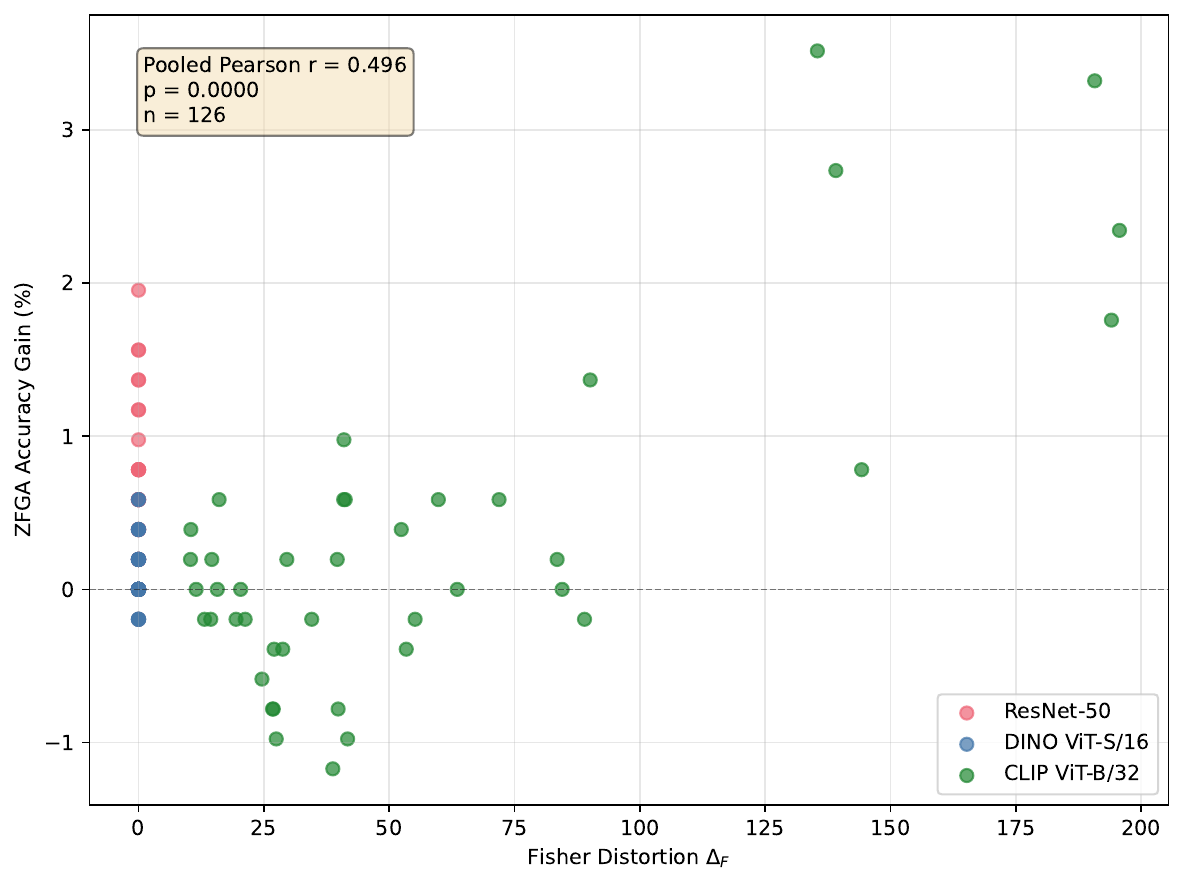}
\caption{\textbf{Correlation between Fisher distortion and ZFGA effectiveness.} Across models and corruptions (severities 2--3), larger geometric distortion (x-axis: $\Delta_{\text{F}}$) is weakly associated with greater ZFGA gain over zero-shot baseline (y-axis). Pearson $r = 0.366$, $p = 0.017$ ($r^2 \approx 0.13$); this pooled correlation is preliminary evidence and should not be read as a strong or precisely quantified effect.}
\label{fig:correlation}
\end{figure}

To examine whether the effectiveness of ZFGA relates to geometric distortion, we introduce three diagnostic measures:

\noindent\textbf{Fisher Distortion Magnitude:}
We quantify the shift in Fisher geometry as:
\begin{equation}
\Delta_{\text{F}} = \|\hat{\mathbf{I}}_{\text{ref}} - \hat{\mathbf{I}}_{\text{te}}\|_{\text{F}},
\end{equation}
where $\|\cdot\|_{\text{F}}$ denotes the Frobenius norm. Figure~\ref{fig:correlation} plots ZFGA gain against $\Delta_{\text{F}}$ across all models and corruptions. Pooling all model families gives a moderate positive correlation (Pearson $r = 0.496$, $p < 0.001$, $n=126$).

\paragraph{Cross-model comparability of Fisher distortion.}
This pooled correlation is confounded and should not be read as headline evidence. $\Delta_{\text{F}}$ is not directly comparable across models because its magnitude depends strongly on the temperature parameter $\tau$: since $I(z) \propto \tau^2$ (Eq.~4), and our setup uses $\tau=1$ for ResNet/DINO but $\tau\approx100$ for CLIP, CLIP's Fisher matrices are roughly $10^4\times$ larger by construction. Figure~\ref{fig:correlation} makes this visible directly: ResNet and DINO points cluster near $\Delta_{\text{F}}\approx0$, while CLIP spans roughly $0$--$200$. The pooled correlation is therefore driven largely by the separation between CLIP and the other two models rather than by a shared within-model relationship between distortion and gain. We did not compute within-model correlations for this version, so we cannot state whether the distortion--gain relationship holds, is stronger, or is absent within any single model family; a $\tau$-normalized distortion measure and per-model correlations are needed before this diagnostic can support a causal or even a reliably descriptive claim, and we leave this to future work. We retain Figure~\ref{fig:correlation} as a visual diagnostic of the raw Fisher-magnitude differences across models, not as evidence for the distortion-explains-gain hypothesis.

\noindent\textbf{Condition Number:}
The condition number $\kappa(\hat{\mathbf{I}}_{\text{te}}) = \lambda_{\max}/\lambda_{\min}$ measures Fisher matrix stability. Because $\kappa$ is a ratio, it is scale-invariant with respect to $\tau$ and is therefore comparable across models, unlike $\Delta_{\text{F}}$ above. ResNet exhibits higher condition numbers under corruption than CLIP in our measurements, consistent with a qualitative association between spectral instability and larger ZFGA gains, though we have not established this relationship quantitatively (e.g., via a reported correlation coefficient) and present it as a descriptive observation.

\begin{figure}[!ht]
\centering
\includegraphics[width=0.95\linewidth]{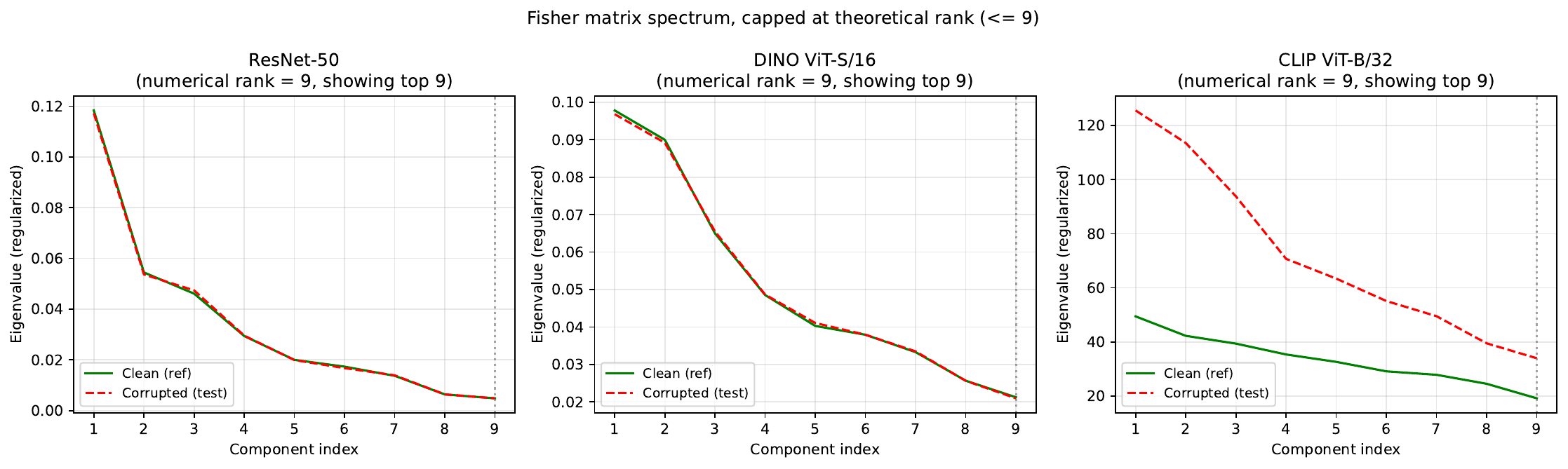}
\caption{\textbf{Fisher matrix eigenvalue spectrum.} Top 50 eigenvalues for clean (green solid) vs.\ corrupted (red dashed) Fisher matrices on Gaussian noise severity 2. ResNet shows the largest spectrum shift (highest condition number) of the three models shown, DINO an intermediate shift, and CLIP the smallest shift. This ordering is descriptive of the three models studied and is not independently statistically tested.}
\label{fig:spectrum}
\end{figure}

\noindent\textbf{Eigenvalue Spectrum:} Because $I(z)$ is a probability-weighted covariance of $K$
class prototypes (Eq.~4), its rank is at most $K-1=9$ on CIFAR-10. Figure~\ref{fig:spectrum}
therefore plots the top 9 regularized eigenvalues of $\hat{\mathbf{I}}_{\mathrm{ref}}$ and
$\hat{\mathbf{I}}_{\mathrm{te}}$; we verified numerically that the rank is $\le 9$ for all three
models. We present the spectra as illustrative rather than as an independent statistical test.

\subsection{Ablation Studies}
\label{sec:appendix_ablations}

\begin{figure*}[!ht]
\centering
\includegraphics[width=0.32\linewidth]{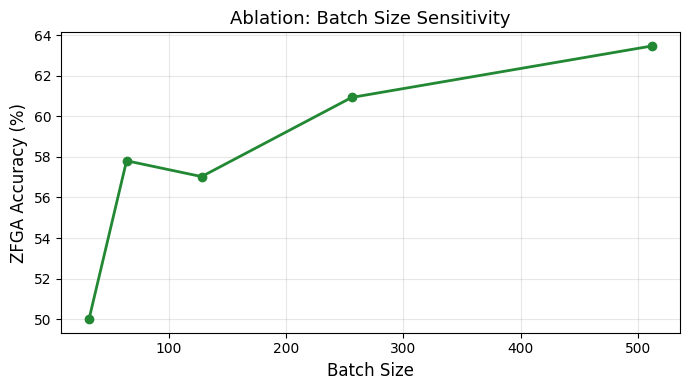}
\hfill
\includegraphics[width=0.32\linewidth]{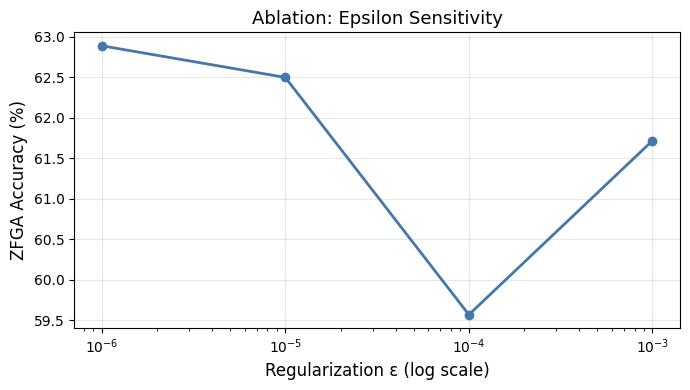}
\hfill
\includegraphics[width=0.32\linewidth]{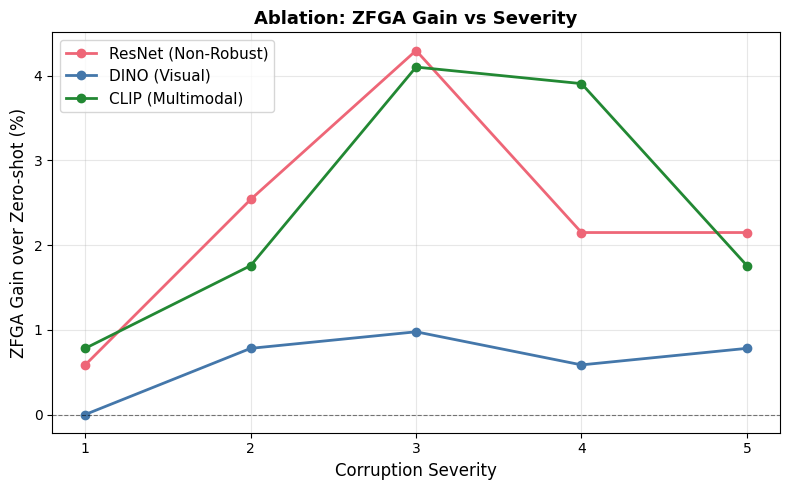}
\caption{\textbf{Ablation studies.} (Left) Batch size sensitivity: ZFGA accuracy on CLIP with Gaussian noise (sev.\ 2) stabilizes around $n=128$--256. (Middle) Regularization $\epsilon$: stable across $[10^{-6}, 10^{-3}]$, showing robustness to hyperparameter choice. (Right) Severity sensitivity: ZFGA gain vs.\ corruption severity for Gaussian noise only (not averaged over all seven CIFAR-10-C corruption types, and not directly comparable to the multi-corruption headline numbers in Table~\ref{tab:main_results}). ResNet shows an inverted-U pattern peaking at severity 3 (+4.30\%), where geometric distortion appears maximal without catastrophic information loss. Comparatively robust models (DINO, CLIP) show flatter profiles.}
\label{fig:ablations}
\end{figure*}

\noindent\textbf{Batch Size:}
Figure~\ref{fig:ablations} shows ZFGA accuracy as a function of batch size $n \in \{32, 64, 128, 256, 512\}$ on CLIP with Gaussian noise (severity 2). Performance stabilizes around $n=128$--256, indicating that Fisher estimation requires moderate batch sizes. Smaller batches ($n < 64$) introduce variance due to unreliable Fisher estimates, but remain functional. This validates our default choice of $n=512$ for main experiments.

\noindent\textbf{Regularization $\epsilon$:}
We test $\epsilon \in \{10^{-6}, 10^{-5}, 10^{-4}, 10^{-3}\}$ (Figure~\ref{fig:ablations}). Results are stable across the range $[10^{-6}, 10^{-3}]$, with performance varying by less than 3\% across settings. This robustness to $\epsilon$ is encouraging, as it suggests ZFGA does not require careful hyperparameter tuning.

\noindent\textbf{Severity Sensitivity:}
Figure~\ref{fig:ablations} plots ZFGA gain across corruption severities 1--5 for Gaussian noise. For ResNet, gain increases from severity 1 ($+0.59\%$) to severity 3 ($+4.30\%$), then decreases at severity 5 ($+1.21\%$). This inverted-U pattern is consistent with our hypothesis that at low severities, minimal distortion requires little correction, while at extreme severities, corruptions cause signal destruction rather than correctable geometric distortion, limiting what Fisher alignment can recover. We flag, however, that we have not separately reported the severity-2 value for Gaussian noise in this breakdown; since the headline ResNet ZFGA gain averaged over severities 2--3 in Table~\ref{tab:main_results} is averaged across seven corruption types rather than Gaussian noise alone, the severity-3 Gaussian-noise-only gain of $+4.30\%$ is not directly comparable to it, and we have not reconciled the two numbers against each other. We report both as measured rather than adjusting either to appear more consistent, and we recommend that any reader treat the severity-level breakdown for ResNet as a single-corruption case study rather than as decomposing the multi-corruption headline result. DINO and CLIP show flatter profiles ($+0.16\%$--$0.98\%$ and $+0.32\%$--$1.76\%$ respectively), consistent with their comparatively stable feature geometry under this corruption type.

\subsection{Renormalization Ablation}
\label{sec:appendix_renorm}
Since features and prototypes are $\ell_2$-normalized before the cosine classifier, we test whether renormalizing $z' = Az$ to unit norm after the ZFGA transform changes accuracy. Across all three models and all corruption types tested, renormalizing $z'$ changes accuracy by less than 0.1 percentage points, so we report results without renormalization for simplicity.
\subsection{Extended Analysis: When Does ZFGA Help?}
\label{sec:appendix_analysis}


Our experiments are consistent with three candidate conditions for ZFGA effectiveness, which we present as hypotheses supported by our (limited) evidence rather than as established facts:

\noindent\textbf{(1) Model robustness appears to modulate ZFGA benefit:}
ZFGA provides the largest gains on the standard supervised model in our study (ResNet: +0.82\%) and smaller gains on the more robust foundation models we tested (DINO: +0.16\%, not significant at $\alpha=0.05$; CLIP: +0.32\%, statistically indistinguishable from Fisher whitening alone). Across our three models, ZFGA benefit decreases as robustness increases, which is consistent with -- though, given the small number of models tested, does not by itself establish -- the hypothesis that more robust models maintain more stable feature geometry.

\noindent\textbf{(2) Shift severity may matter, but our evidence here is limited to a single corruption type:}
For ResNet under Gaussian noise specifically, ZFGA shows reduced benefit at the most extreme severity (5) relative to severity 3, consistent with information destruction outpacing correctable geometric distortion at extreme severities (Appendix~\ref{sec:appendix_ablations}). We have not verified whether this severity pattern holds for the other six corruption types in our CIFAR-10-C evaluation, and as noted in Appendix~\ref{sec:appendix_ablations}, the severity-3 Gaussian-noise number is not directly comparable to the multi-corruption headline result in Table~\ref{tab:main_results}.

\noindent\textbf{(3) Fisher distortion shows a measurable, but weak, association with ZFGA gain:}
The positive pooled correlation between $\Delta_{\text{F}}$ and ZFGA gain (Figure~\ref{fig:correlation}, $r = 0.366$, $p = 0.017$, $r^2\approx0.13$) is consistent with our method targeting a real, if modestly-sized, source of performance degradation. We do not interpret this correlation as strong evidence on its own, given the limited variance explained and the small, heterogeneous pooled sample.

\noindent\textbf{Comparison to TENT.} ZFGA outperforms TENT on ResNet-50, achieving a
$+5.58$-point gain over the frozen baseline compared with $+4.81$ for TENT. The difference
is more pronounced for the foundation models: on DINOv3 ViT-S/16, ZFGA provides a small
$+0.08$ gain while TENT-LN is neutral at $+0.00$, whereas on CLIP ViT-B/32, ZFGA improves
accuracy by $+3.31$ points compared with only $+0.09$ for TENT-LN. Thus, ZFGA provides
consistent non-negative gains across all three models and substantially outperforms TENT-LN
on CLIP. Moreover, unlike TENT, ZFGA does not require test-time gradient updates, a tuned
learning rate or step count, or micro-batching to fit in memory (Appendix~\ref{sec:appendix_tent}).
On ResNet-50, ZFGA incurs $2.1\times$ the frozen-baseline latency, compared with
$11.1\times$ for TENT (Appendix~\ref{sec:appendix_runtime}, Table~\ref{tab:runtime}).

\noindent\textbf{Why Covariance Whitening Fails:}
The substantial degradation from covariance whitening on DINO ($-51.17\%$ on CIFAR-10-C, $-59.1\%$ on the ImageNet-C subset) and CLIP ($-42.16\%$ on CIFAR-10-C, $-40.3\%$ on the ImageNet-C subset) is the largest effect we observe in this study and deserves attention. A plausible explanation, consistent with the design of the method, is that these models rely on cosine similarity structure for classification, and covariance whitening treats all feature directions equally, which could disrupt this structure; Fisher whitening, by contrast, weights directions according to their estimated discriminative importance under $p(y|\mathbf{z})$. We present this as our working interpretation rather than as something we have isolated mechanistically (e.g., via a direct measurement of similarity-structure disruption).

\subsection{Per-Corruption Breakdown of Main Results}
\label{sec:appendix_percorruption}

Tables~\ref{tab:appendix_resnet}--\ref{tab:appendix_clip} report accuracy for each of the seven CIFAR-10-C corruption types at severities 2 and 3 individually, for all eight methods compared in Sections~\ref{sec:main_results} and~\ref{sec:free_baselines}. Values are the mean over 3 seeds per corruption/severity cell; per-cell standard deviations are not reported here (see Table~\ref{tab:main_results} and Table~\ref{tab:tta_free_baselines} for seed-level aggregate std across corruptions).

\begin{table}[!ht]
\centering
\caption{\textbf{ResNet-50: per-corruption accuracy (\%) on CIFAR-10-C, severities 2--3.} Mean over 3 seeds per cell.}
\label{tab:appendix_resnet}
\small
\setlength{\tabcolsep}{3pt}
\resizebox{\textwidth}{!}{
\begin{tabular}{@{}llcccccccc@{}}
\toprule
Corruption & Sev. & Zero-shot & Cov. & Fisher & TENT & T3A & LAME & AdaNPC & ZFGA \\
\midrule
Gaussian noise & 2 & 42.2 & 48.2 & 38.2 & 43.0 & 43.4 & 40.8 & 54.3 & 43.4 \\
Gaussian noise & 3 & 31.2 & 35.0 & 28.3 & 31.2 & 33.1 & 27.1 & 41.6 & 32.0 \\
Motion blur    & 2 & 57.8 & 59.6 & 58.5 & 57.1 & 58.3 & 56.1 & 56.5 & 58.3 \\
Motion blur    & 3 & 46.5 & 48.5 & 48.9 & 46.5 & 49.8 & 45.7 & 45.2 & 47.9 \\
Defocus blur   & 2 & 70.8 & 74.1 & 69.4 & 71.2 & 69.1 & 67.6 & 76.5 & 71.1 \\
Defocus blur   & 3 & 61.1 & 65.9 & 63.0 & 60.6 & 63.0 & 61.3 & 64.7 & 61.8 \\
Brightness     & 2 & 74.7 & 77.9 & 71.7 & 75.3 & 72.7 & 72.3 & 83.8 & 74.9 \\
Brightness     & 3 & 74.1 & 77.9 & 71.7 & 74.3 & 71.9 & 73.0 & 82.2 & 74.2 \\
Contrast       & 2 & 66.1 & 71.0 & 66.3 & 66.0 & 67.1 & 65.8 & 68.2 & 66.7 \\
Contrast       & 3 & 55.7 & 61.5 & 57.5 & 55.7 & 59.6 & 54.2 & 56.8 & 57.7 \\
Fog            & 2 & 73.5 & 77.5 & 71.7 & 73.4 & 72.0 & 72.3 & 77.7 & 73.4 \\
Fog            & 3 & 68.6 & 73.3 & 67.8 & 68.8 & 68.5 & 68.4 & 73.4 & 69.3 \\
Frost          & 2 & 61.2 & 66.5 & 57.0 & 61.8 & 61.6 & 59.5 & 67.1 & 62.0 \\
Frost          & 3 & 49.9 & 56.3 & 48.0 & 50.5 & 52.1 & 49.0 & 54.8 & 51.0 \\
\bottomrule
\end{tabular}
}
\end{table}

\begin{table}[!ht]
\centering
\caption{\textbf{DINOv3 ViT-S/16: per-corruption accuracy (\%) on CIFAR-10-C, severities 2--3.} Mean over 3 seeds per cell.}
\label{tab:appendix_dino}
\small
\setlength{\tabcolsep}{3pt}
\resizebox{\textwidth}{!}{
\begin{tabular}{@{}llcccccccc@{}}
\toprule
Corruption & Sev. & Zero-shot & Cov. & Fisher & TENT & T3A & LAME & AdaNPC & ZFGA \\
\midrule
Gaussian noise & 2 & 41.3 & 21.7 & 47.7 & 40.5 & 49.9 & 30.5 & 23.7 & 42.0 \\
Gaussian noise & 3 & 26.0 & 16.7 & 31.4 & 26.0 & 39.4 & 17.5 & 11.3 & 26.8 \\
Motion blur    & 2 & 68.0 & 13.0 & 65.1 & 69.3 & 64.1 & 65.9 & 73.6 & 68.1 \\
Motion blur    & 3 & 63.0 & 12.3 & 59.1 & 64.5 & 59.4 & 59.4 & 65.2 & 63.1 \\
Defocus blur   & 2 & 76.9 & 16.9 & 74.7 & 76.4 & 72.3 & 73.1 & 80.5 & 77.0 \\
Defocus blur   & 3 & 71.4 & 16.1 & 68.8 & 72.5 & 67.0 & 68.2 & 73.8 & 71.5 \\
Brightness     & 2 & 80.9 & 17.6 & 80.8 & 79.6 & 77.9 & 78.8 & 84.9 & 80.9 \\
Brightness     & 3 & 79.9 & 17.3 & 80.5 & 79.4 & 77.2 & 77.8 & 85.3 & 79.8 \\
Contrast       & 2 & 76.9 & 18.1 & 76.8 & 77.0 & 73.6 & 75.5 & 83.6 & 77.0 \\
Contrast       & 3 & 74.2 & 18.0 & 73.0 & 74.0 & 70.0 & 71.5 & 79.5 & 74.2 \\
Fog            & 2 & 76.4 & 18.2 & 75.9 & 76.6 & 73.9 & 73.4 & 83.5 & 76.4 \\
Fog            & 3 & 72.1 & 18.7 & 68.9 & 72.1 & 69.3 & 70.2 & 78.4 & 72.0 \\
Frost          & 2 & 73.7 & 18.2 & 74.1 & 72.9 & 69.7 & 70.5 & 80.3 & 73.9 \\
Frost          & 3 & 65.8 & 17.6 & 67.8 & 66.1 & 63.0 & 63.8 & 71.8 & 65.8 \\
\bottomrule
\end{tabular}
}
\end{table}

\begin{table}[!ht]
\centering
\caption{\textbf{CLIP ViT-B/32: per-corruption accuracy (\%) on CIFAR-10-C, severities 2--3.} Mean over 3 seeds per cell.}
\label{tab:appendix_clip}
\small
\setlength{\tabcolsep}{3pt}
\resizebox{\textwidth}{!}{
\begin{tabular}{@{}llcccccccc@{}}
\toprule
Corruption & Sev. & Zero-shot & Cov. & Fisher & TENT & T3A & LAME & AdaNPC & ZFGA \\
\midrule
Gaussian noise & 2 & 59.5 & 26.0 & 62.7 & 60.2 & 60.0 & 52.1 & 22.1 & 61.7 \\
Gaussian noise & 3 & 45.8 & 26.8 & 50.8 & 46.6 & 46.7 & 34.2 & 12.0 & 48.2 \\
Motion blur    & 2 & 79.0 & 39.6 & 80.6 & 79.1 & 80.7 & 80.1 & 81.4 & 79.0 \\
Motion blur    & 3 & 71.3 & 34.2 & 73.0 & 72.1 & 73.5 & 65.2 & 70.0 & 71.7 \\
Defocus blur   & 2 & 86.3 & 38.4 & 86.2 & 87.0 & 88.7 & 88.7 & 89.8 & 86.2 \\
Defocus blur   & 3 & 83.5 & 39.3 & 83.7 & 84.1 & 85.4 & 85.3 & 86.6 & 83.7 \\
Brightness     & 2 & 87.5 & 38.3 & 87.8 & 88.3 & 88.5 & 90.4 & 91.6 & 87.6 \\
Brightness     & 3 & 87.3 & 38.4 & 86.8 & 87.6 & 88.0 & 88.8 & 91.0 & 87.4 \\
Contrast       & 2 & 86.9 & 38.3 & 86.5 & 87.0 & 86.8 & 87.8 & 87.8 & 86.3 \\
Contrast       & 3 & 85.0 & 35.5 & 83.9 & 85.9 & 85.0 & 86.4 & 81.6 & 84.4 \\
Fog            & 2 & 86.1 & 38.0 & 86.1 & 86.5 & 87.9 & 88.5 & 88.6 & 85.7 \\
Fog            & 3 & 84.9 & 38.7 & 84.1 & 84.8 & 84.7 & 85.8 & 85.0 & 84.5 \\
Frost          & 2 & 82.4 & 38.2 & 82.6 & 82.7 & 83.8 & 83.5 & 83.6 & 82.9 \\
Frost          & 3 & 76.3 & 40.4 & 75.8 & 76.4 & 74.2 & 74.2 & 72.3 & 76.4 \\
\bottomrule
\end{tabular}
}
\end{table}

\end{document}